\documentclass{IOS-Book-Article}

\usepackage{mathptmx}
\usepackage{soul}\setuldepth{article}
\usepackage{booktabs} 
\usepackage{multirow} 
\usepackage{caption}   
\usepackage{graphicx}  
\usepackage{float}
\usepackage{todonotes}
\usepackage{enumitem}

\def\hb{\hbox to 11.5 cm{}}

\begin{document}

\pagestyle{headings}
\def\thepage{}
\begin{frontmatter}              

\title{Using LLMs to Discover Legal Factors} 

\markboth{}{October 2024\hb}

\markboth{}{December 2023\hb}
\author[A]{\fnms{Morgan} \snm{Gray}\orcid{0000-0002-3800-2103}
\thanks{Corresponding Author: Morgan Gray, Learning Research and Development Center: 3420 Forbes Ave. Pittsburgh, PA 15260, USA; Email: mag454@pitt.edu}},
\author[C]{\fnms{Jaromir} \snm{SAVELKA}\orcid{0000-0002-3674-5456}}
\author[D]{\fnms{Wesley} \snm{OLIVER}\orcid{0000-0002-3873-8479}}
and
\author[A,B]{\fnms{Kevin} \snm{ASHLEY}\orcid{0000-0002-3873-8479}}

\runningauthor{{M.A. Gray, et al.}}
\address[A]{Intelligent Systems Program, University of Pittsburgh, USA}
\address[B]{School of Law, University of Pittsburgh, USA}
\address[C]{School of Computer Science, Carnegie Mellon University, USA}
\address[D]{School of Law, Duquesne University, USA}

\begin{abstract}
Factors are a foundational component of legal analysis and computational models of legal reasoning.  These factor-based representations enable lawyers, judges, and AI and Law researchers to reason about legal cases.
In this paper, we introduce a methodology that leverages large language models (LLMs) to discover lists of factors that
effectively represent a legal domain.  
Our method takes as input raw court opinions and produces a set of factors and associated definitions. We demonstrate that a semi-automated approach, incorporating minimal human involvement, produces factor representations that can predict case outcomes with moderate success, if not yet as well as expert-defined factors can.

\end{abstract}

\end{frontmatter}
\markboth{October 2024\hb}{October 2024\hb}

\section{Introduction}

Recently, large language models (LLMs) have been applied automatically to annotate legal case texts from particular legal domains in terms of factors from pre-existing factor lists. In this paper, we describe and assess a methodology for employing LLMs to discover factors in case texts without using a pre-existing factor list. 

Our method takes as input raw court opinions and produces a set of factors and associated definitions. We evaluate the extent to which an LLM can identify from scratch {\it any} factors in the cases from a legal domain where the LLM has no apparent access to a pre-existing list of factors or their definitions for that domain.  We demonstrate that a semi-automated approach, with a human in the loop produces factor representations that can predict case outcomes with moderate success, if not yet as well as expert-defined factors can.  In the absence of predefined factors from courts or legislative bodies, legal scholars manually analyze hundreds of cases to identify factors, a process that is highly time-consuming and costly.  Our methodology could enable a more efficient process of identifying factor representations of legal domain cases.

Our novel methodology relies on LLMs to perform a task similar to that which lawyer readers, as distinguished from judges, perform in identifying new factors in case texts.  When a judge identifies a new factor in a case for decision, we assume that the judge applies legal knowledge about the type of claim and about relevant cases to induce from the facts of the case at hand a generalized description of a relevant pattern, that is, a factor, that the judge believes she should take into account in deciding. Typically, the judge reports the factual pattern in the findings of fact and indicates that she is deciding the case because of (or despite) this finding. By contrast, in reading the published case text, human readers may simply note the finding of fact and identify it as an instance of a fact-based reason that may apply in other similar cases in the legal domain. Our prompts instruct LLMs to perform a task similar to the reader’s task.

While this paper provides a proof-of-concept, our methodology could eventually be useful in a number of applications. For instance, much work in empirical legal studies and in AI and Law employs pre-defined lists of factors in statistical or machine learning models of case-based legal reasoning. Meanwhile, courts are identifying novel factors in the cases they decide. Our methodology could help to automate the process of detecting new factors to incorporate into subsequent models.  In fact, an LLM in our experiment did discover a new factor or sub-factor potentially augmenting an existing factor list.

\section{Related Work}
Legal factors have played an important role both in empirical legal studies and in AI and Law. Factors are ``consideration[s] a decision maker must or may take into account to determine an outcome.’’ ``[They] can be prescribed in a statute or regulation, or created by courts,'' \cite[p. 2, 3]{rempell2022factors}. Courts employ factors in a variety of legal domains, such as assessing spousal support or determining violations of the right to a speedy trial  \cite[p. 2f]{rempell2022factors},  determining copyright fair use, works made for hire, or consumer confusion as to the source of goods in trademark infringement, (Beebe \cite[p. 1584f]{beebe2006empirical}, \cite{beebe2007empirical,beebe2020empirical}). Empirical legal scholars have studied courts’ use of multi-factor tests in legal domains. They often begin with a canonical list of factors as set out in a statute or in appellate court decisions and apply machine learning (e.g., decision trees)  to evaluate which factors are most important and how courts employ them \cite{beebe2006empirical,beebe2007empirical,beebe2020empirical,shao2022factors}.

AI and Law researchers have computationally modeled legal case-based reasoning and argumentation in terms of  factors. For purposes of these models, factors have been defined operationally as stereotypical fact patterns that tend to strengthen or weaken a side's argument in favor of a legal claim \cite{ashley1990modeling,bench2017hypo,grabmair2016modeling,chorley2005empirical,chorley2005agatha,westermann2019using}.  Factors have been used to model shifts in legal concepts as applied over time in legal decisions \cite{rissland1995detecting}. More recent models employ factors in developing formal models of precedential constraint \cite{horty2012factor}.

Traditionally, representing cases in terms of factors has been a manual activity, whether for statistical analysis or computational modeling. Researchers  read  cases and identified the sentences from which one could infer that facts associated with a factor had occurred in the cases and indicating that a court decided the case as it did because, or in spite, of the presence of that factor.  See, e.g., \cite{gray2022toward} More recently, researchers have developed machine learning pipelines automating to some degree annotating factors in case-related texts, for example trade secrets factors in law-student-prepared case summaries by Ashley and Br{\"uninghaus \cite{ashley2009automatically} or case opinions by Falakmasir and Ashley \cite{falakmasir2017utilizing}, factor values in divorce cases by Li, et al. \cite{li2019research}, factor findings related to WIPO (World Intellectual Property Organization) domain name dispute cases by Branting, et al.\cite{branting2021scalable} such as PriorBizUse (i.e., ``Bona fide business use of Domain Name or demonstrable preparations to do so, prior to notice of the dispute’’), or factors of suspicion in drug interdiction auto stop cases by Gray, et al.\cite{gray2022toward,gray2023can,gray2023automatic}. These automated factor annotation methods could enable researchers who perform empirical legal studies or build computational models of legal argument to automatically classify factors in much larger numbers of cases, potentially increasing their works’ accuracy and scope. In this work, we take the process further by inducing factors from case opinions without the need for a pre-defined list (akin to themes in \cite{drapal2023using}).

AI and Law work on case-based reasoning usually assumes that the legal domain modeled is static as is the canonical list of factors. Empirical legal scholars often assume a static list of factors for their statistical analyses or decision trees. Employing canonical lists of factors, whether with manual or automated annotation, however, begs the question of how to identify and incorporate new factors. Courts do create and apply additional factors even if infrequently \cite[p. 1594, 1646]{beebe2006empirical}. Given the ``open-ended language” of some listed statutory factors, sub-factors occasionally emerge as courts flesh out their meaning in specific cases \cite[p. 610]{beebe2007empirical}. 
As social values or technology change, courts identify new patterns of case facts whose effect on a decision should be taken into account.  According to Rempell \cite[p. 48]{rempell2022factors}, 
``When additionally relevant factors do emerge, often they are incorporated into the prospective list that subsequent decisions will review.''

\subsection{Research Questions}

In light of the related work, we evaluate the following open research questions: 

\begin{enumerate}
\item[RQ1:] How effective are LLMs at automatically synthesizing a factor-based representation of a legal domain from raw court opinions?
\item[RQ2:] How effective are humans at synthesizing a factor-based representation of a legal domain from raw court opinions with the assistance of LLMs?
\item[RQ3:] How effectively can LLMs discover new factors or propose meaningful variations to existing factors within a pre-defined list?
\end{enumerate}

\section{Data}

Our raw data come from the DIAS Corpus, obtained through the Harvard Caselaw Access Project (CAP), \cite{gray2022toward, gray2023automatic}, which contains U.S. court opinions on whether police officers had reasonable suspicion of drug trafficking.  The cases were identified through the efforts of lawyers and law students in \cite{gray2022toward, gray2023automatic}. Out of the corpus of 300, we work with a random subset of 174 cases.  Random sampling ensures unbiased selection, and the smaller subset preserves data for future testing.  The Fourth Amendment governs the admissibility of evidence in federal and state courts and state courts follow the guidance of the US Supreme Court in their interpretation of the amendment. Since every decision — in a federal or state court — interpreting reasonable suspicion in a drug interdiction stop relies on the same legal standard, our data can come from any jurisdiction in the U.S. 

\section{Methodology}

The first step is to process raw opinions by identifying the court's analysis and conclusion sections. This reduces both data noise and the amount of text needed for prompts in later stages.
While courts may present  relevant facts throughout the opinion, in the analysis section, they consistently highlight factors that are relevant to their conclusions. To this end, we prompted \texttt{gpt-4-1106-preview} to identify the analysis portion of an opinion as well as the court's conclusion on the issue of whether the officer had reasonable suspicion to make the detention.  Essentially, we sought to replicate a portion of the work in \cite{savelka2021lex,savelka2018segmenting} with zero-shot LLMs.  We numbered each paragraph and instructed the model with a prompt providing: 1) Explanation of what an analysis section of an opinion is; 2) explanation of what a court's conclusion is; 3) an example of a court's analysis on the relevant issue; 4) an example of the court's conclusion on the relevant issue; and 5) instructions to return the span of paragraphs encompassing the court's analysis and conclusion.



To assess the model’s performance, we annotated a sub-sample of 103 out of 173 cases used in this work. An attorney, an expert in drug interdiction law, identified the paragraphs containing the court's analysis and conclusion. We then collected the sets of factors identified through gold standard annotation that describe factors of suspicion in the text.  We focus on identifying factor sets as they will be converted into a dichotomous vector for use in evaluation, as explained in Section \ref{sec:eval}. Table \ref{tab:spans} shows the results of comparing the sets of gold standard factors identified by the model and annotator. 
For each of the 103 examples, individual scores for recall, precision, and accuracy were calculated, and the average of these scores across all cases is reported in the table.  
Recall measures the extent to which the model identified the same set of factors in the analysis and conclusion spans as the expert annotator. Here, we focus on recall to ensure that factors are not being missed. The high recall score indicates the process is reliable.

\begin{table}[]
\centering
\begin{tabular}{|l|l|l|l|}
\hline
\textbf{Average}     & \textbf{Accuracy} & \textbf{Precision} & \textbf{Recall} \\ \hline
\textbf{Value}      & 0.96              & 0.99               & 0.97            \\ \hline
\end{tabular}
\caption{Analysis of factor sets between text spans identified by \texttt{gpt4-turbo} and gold standard annotations.}
\label{tab:spans}
\end{table}

\subsection{Inducing Factors with LLMs}
\label{sec:inducing_rough}

In the next stage, we prompt an LLM to replicate the work of an attorney in reviewing numerous case opinions, identifying relevant facts and potential factors, and ultimately defining a canonical list of factors.
We structured a prompt to focus only on the legal issue at hand: ``the legal issue here is whether a police officer had reasonable suspicion to detain a motorist.''  The prompt contains no other knowledge about the domain or issue.  We then introduce the concept and definition of a legal factor.  Next, we provide an example of a factor from another legal domain, namely  
the example of the ``Disclosure-in-negotiations'' factor from the trade-secret domain \cite{ashley2017artificial}: 

\begin{quote}
This factor describes a plaintiff's disclosure of its product information in negotiations with a defendant. You should notice how this description of the factor is sort of broad, but specific. In specific, it is referencing about the disclosure of facts during negotiations. Broadly, a disclosure could happen in negotiation in many different ways. Perhaps the disclosure occurred during informal negotiations during a dinner meeting. Perhaps the disclosure occurred during a formal negotiation session.
\end{quote}

Next, we instruct the model that a factor must directly support a court's conclusion on the legal issue, and give the example of a disclosure-in-negotiations as supporting a court's conclusion.  Lastly, we instruct the model to search for factors based on the issue of reasonable suspicion to detain a motorist.  

We used two different models: \texttt{gpt-4-turbo} and \texttt{llama-3.1-70b-versatile}.  Both models provide a context window of up to or greater than  128,000  tokens. The inference process for both models used the parameters \texttt{Temperature=0}, \texttt{Top P=1}, \texttt{Max Tokens=4000}, \texttt{Frequency Penalty=0}, and \texttt{Presence Penalty=0}.  We rely on these settings to produce as deterministic and repeatable output as possible.   To induce factors, we randomly selected 50 cases and automatically extracted the court's analysis and conclusion sections using \texttt{gpt-4-turbo}. 

We derived a rudimentary set of factors by dividing the 50 cases into 5 sets of 10 and prompting the model as described above. We chose this approach due to concerns about model performance. As noted in \cite{jiang2024longllmlinguaacceleratingenhancingllms}, large context windows can result in reduced sensitivity to information in the middle of lengthy inputs. To mitigate this, we prompted the model in smaller batches to ensure it could process all information with adequate sensitivity. Using smaller sets of cases also improves explainability, as it is easier for a human to review 10 cases at a time rather than 50 when scrutinizing the LLM’s output.  

In this way,  \texttt{gpt-4-turbo}, for example, produced a rough set of factors, such as:
\begin{quote}
{\it Physical Indicators of Stress}: Physical indicators of stress, such as shaking or excessive sweating, beyond what might be expected during a routine traffic stop.\\
{\it Driver's Behavior and Statements}: Observable nervousness beyond what is typical, inconsistent or implausible explanations for travel, and changes in the driver's story can contribute to reasonable suspicion.
\end{quote}

\subsection{Refined Factor Representation of Rudimentary Induced Factors}
\label{sec:refining}

Next, the rough factors are refined into a Refined Factor Representation (RFR). The goal is to define each factor specifically enough to capture its core concept, while keeping it broad enough to cover all its potential manifestations.
For example, a factor such as the one induced in \ref{sec:inducing_rough} is refined into something like the following:

\begin{quote}
Nervous Behavior and Evasive Answers:
Exhibiting nervous behavior such as excessive sweating, avoiding eye contact, rapid breathing, or shaking, and providing evasive or inconsistent answers during a traffic stop can contribute to an officer's reasonable suspicion of criminal activity. This includes observable nervousness beyond what is typical, changes in the driver's story, and actions such as avoiding eye contact or attempting to distance oneself from the vehicle.
\end{quote}

 To accomplish this, we asked a human annotator to perform the refinements following a set of guidelines. The annotator was blind in the sense that he was not previously aware of the reasonable suspicion for drug trafficking domain or  the factors from our prior work \cite{gray2022toward, gray2023automatic}. He read three drug trafficking cases and the guidelines, which explained factor-based legal domains using an example from trade-secret misappropriation and included the LLM prompt that produced the rough factors described in Section \ref{sec:inducing_rough}.  The annotator was then instructed to refine each of the rough factors as per the guidelines:

\begin{quote}
The key is to identify factors in such a manner that they have meaningful commonality, but meaningful differences from other factors.  Meaning there should be a definable boundary between factor definitions, but not so highly defined that each fact that you read is designated as its own factor .... It is key to understand how one kind of factor can be described in many different ways. 
\end{quote}
Lastly, the annotator was instructed to: 

\begin{enumerate}[noitemsep,topsep=1.5pt]
    \item combine obviously identical or highly similar induced factors identified across prompt iterations,  
    \item identify subtle similarities and differences, and group or distinguish factors accordingly,  
    \item refine redundant language, 
    \item capture the full breadth of a factor, 
    \item avoid combining what a human may identify as separate factors under a single definition,
    \item consider counting as a factor something for which the LLM output contained only a single instance, since the LLM has seen only a relatively small sample of cases.
\end{enumerate}

In a separate activity, we instructed an LLM to perform the same factor refinement task using a prompt that was identical to the factor refinement guidelines given to the human annotator.   We relied on two large context window models \texttt{gpt-4-turbo} and \texttt{llama-3.1-70b-versatile}. Each model produced a refined factor representation (RFR) of a legal domain based on the raw induced factors. 

\section{Evaluation}
\label{sec:eval}



Here, we evaluate the quality of the RFR synthesized from raw LLM-induced factor outputs for the domain of reasonable suspicion of drug trafficking.  Our evaluation assessment is based on the experiment in \cite{gray2023automatic} where sets of factors identified in court opinions were converted into a binary vector representing the individual facts present/absent in a particular case.  In order to identify the factor sets we rely on annotations of factors described in sentences as in \cite{gray2023can}.  Like \cite{gray2023automatic} we use the binary representation to make predictions about whether the court concluded that suspicion was found.  By using this evaluation method, we can measure whether the refined factor representation produces factor sets that represent the domain in such a manner that meaningful predictions and analysis can be made.  We ultimately assess the reliability of the refined factor representations by comparing 4 factor representations that differ based on the annotator that was used to identify factor sets and on the refiner who/that refined the factor representation:

\begin{description}
    \item[Gold/DIAS CFR] gold standard annotations made by an expert annotator using the Canonical Factor Representation (CFR) of DIAS factors identified in \cite{gray2022toward,gray2023can,gray2023automatic}. 
    \item[LLM/DIAS CFR:] annotations made by an LLM (\texttt{gpt-4o} or \texttt{llama3-70b-8192}), using a guideline prompt prepared for human annotator using the Canonical Factor Representation of DIAS factors identified in \cite{gray2022toward,gray2023can,gray2023automatic}.
    \item[LLM/Human RFR:] annotations made by an LLM (\texttt{gpt-4o} or \texttt{llama3-70b-8192}), using human's RFR of raw factors induced by an LLM in \ref{sec:inducing_rough}. 
    \item[LLM/LLM RFR:] annotations made by an LLM (\texttt{gpt-4o} or \texttt{llama3-70b-8192}), using LLM's RFR, using the same instructions as the human in \textbf{[LLM/Human RFR]}, of raw factors induced by an LLM in \ref{sec:inducing_rough}. 
\end{description}

The collection of cases in this evaluation has a prevalent class imbalance.  The court found that reasonable suspicion was present in 71\% of the cases and not present  in 29\% of the cases.  Because of this imbalance, we rely on the Matthews Correlation Coefficient (MCC) to assess model performance.  The MCC measures the correlation between a model's predicted labels and the true labels \cite{totsch2021matthews} and is appropriate where there is a class imbalance.  The MCC returns a value from 1 to -1, where 1 indicates perfect classification, 0 indicates random classification, and -1 indicates total disagreement with the predicted and true labels.  We use two models, ElasticNet and Random Forest, to evaluate the quality of the predictions.  Both models were chosen because of their capability of handling class imbalance -- ElasticNet through its regularization capabilities and Random Forest by using balanced class weighting.  We used an 80-20 training-testing scheme.  Both models were trained using 5 fold cross-validation and were tuned using a grid search, finding the best model parameters based on MCC score.  All models were trained and evaluated on the same training-testing split.   

\begin{table}[]
\centering
\caption{MCC Scores Across Different Models and Settings}
\label{tab:models_settings}
\begin{tabular}{@{}lccccccc@{}}
\toprule
\multirow{2}{*}{Model} & \multicolumn{2}{c}{LLM/Human RFR} & \multicolumn{2}{c}{LLM/DIAS CFR} & \multicolumn{2}{c}{LLM/LLM RFR} & \multicolumn{1}{c}{Gold/DIAS CFR} \\
    & llama     & gp4-o    & llama    & gp4-o   & llama    & gp4-o   &        \\
    & MCC  & MCC  & MCC & MCC & MCC & MCC & MCC \\ \midrule
ElasticNet           &0.56$+$ &0.30$\triangle$ &0.52$+$ &0.32$\triangle$ &0.24$\circ$ &0.36$\triangle$ & 0.56$+$  \\
RandomForest         &0.46$+$ &0.31$\triangle$ &0.61$+$ &0.41$+$         &0.08$-$     &0.20$\circ$     & 0.66$+$   \\
Majority Class          &0.00,0.72 & & & & & &   \\
Random Label         &-0.07,0.44 & & & & & &   \\
\bottomrule
\\
\end{tabular}
\captionsetup{labelformat=empty}
\caption*{
\makebox[\linewidth]{
    \parbox{0.95\linewidth}{\textbf{Strength of MCC:} $\star$ Very Strong (0.9--0.7), $+$ Strong (0.4--0.6), $\triangle$ Moderate (0.3), $\circ$ Weak (0.2), $\diamond$ Negligible (0.1), $-$ None}
}}
\captionsetup{labelformat=default}

\end{table}

We note that in the case of binary classification, the MCC is identical to the Phi($\phi$) correlation statistic \cite{chicco2021matthews} and the Pearson correlation coefficient for two binary variables and is, thus, similar in interpretation \cite{godino2018hunter}.  For this reason, we rely on the familiar thresholds for interpreting the Pearson correlation coefficient.  We interpret positive MCC values according to the following thresholds: very strong (0.9-0.7), strong (0.4-0.6), moderate (0.3), weak (0.2), negligible (0.1) \cite{haldun2018users}.

When we examine the predictions in Table~\ref{tab:models_settings} from the gold standard annotations, we see a strong correlation between the models' predictions and the true labels.  
When using the factor sets identified by \texttt{gpt-4o} and \texttt{llama3-70b-8192} using the DIAS Canonical Factor Representation (CFR), we see a strong performance from both \texttt{gpt-4o} and \texttt{llama3-70b-8192}, with \texttt{llama3-70b-8192} clearly out performing \texttt{gpt-4o}.  Notably, the predictions with the factor sets identified by \texttt{llama3-70b-8192} using the DIAS representation were very close in  quality to the gold standard predictions. This suggests that the CFR of DIAS factors is robust across annotators.  

The Refined Factor Representation produced by the human annotator (LLM/Human) produced strong predictions with \texttt{llama3-70b-8192} and moderate predictions with \texttt{gpt-4o}.  Particularly, annotations made in accordance with the LLM/Human Refined Factor Representation using \texttt{llama3-70b-8192} were similar in performance to gold standard annotations.  The lowest overall performing group were the annotations made with an LLM using Refined Factor Representations identified by \texttt{gpt-4o} and \texttt{llama3-70b-8192}.  Ultimately these results suggest that an entirely automated pipeline from raw opinions to a Refined Factor Representation would need improvement. 

\section{Discussion}
\begin{figure}[]
    \centering
    \includegraphics[width=0.95\textwidth, height=0.4\textheight]{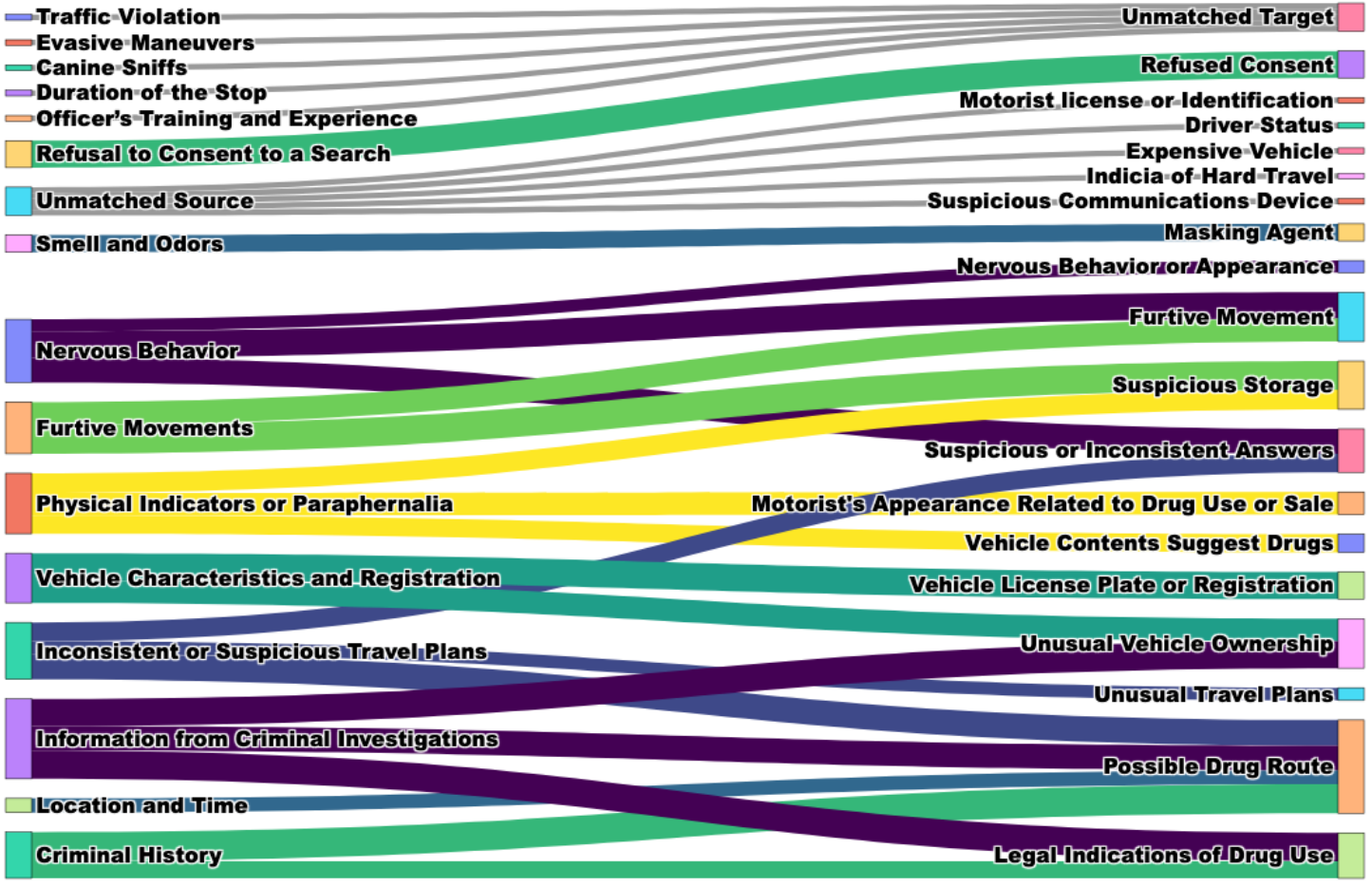}
    \caption{Similarity between Human Refined Factor Representation (left) and DIAS Canonical Factor Representation (right).}
    \label{fig:human-dias}
\end{figure}

The following diagrams help to explain the difference in predictive performance of factors identified in the Dias CFR and those in the RFR synthesized by the human annotator (Human RFR) or LLMs (Llama RFR or GPT RFR).

The diagrams depict  semantic relations between these two sets of factors. We embedded the definitions of the factor representations and then measured the cosine similarity between the RFR and CFR factors. For each RFR factor, we calculated the three most similar CFR factors. To filter out low similarity  matches, we disregarded any similarity score within the top three for an RFR-CFR factor pair that was lower than the average of all top three scores. The plot in  Figure \ref{fig:human-dias}  demonstrates the similarity between Human RFR factors on the left and CFR factors on the right.  The plot in  Figure \ref{fig:llama-dias}  demonstrates the similarity between Llama RFR factors on the left and CFR factors on the right. If the CFR factor is connected to an ``Unmatched Source'' it means that no similar  RFR factor was identified.  If an RFR Factor on the left is connected to ``Unmatched Target'' it means that the  RFR factor was not identified as similar to any CFR factor.  The weight of the line indicates the strength of the similarity; heavier lines signify higher cosine similarity. 

\begin{figure}[]
    \centering
    \includegraphics[width=0.95\textwidth, height=0.4\textheight]{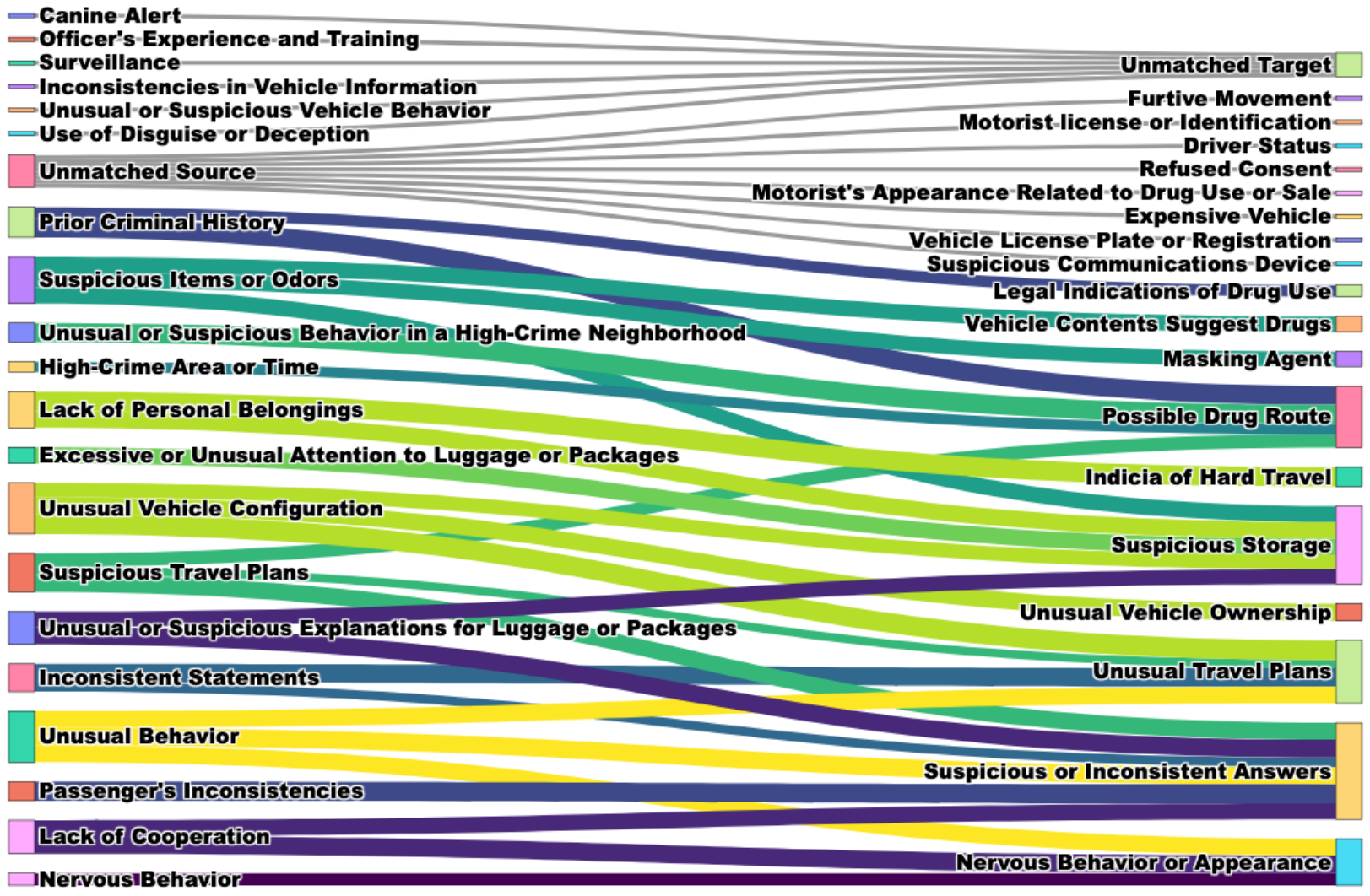}
    \caption{Similarity between Llama Refined Factor Representation (left) and DIAS Canonical Factor Representation (right).}
    \label{fig:llama-dias}
\end{figure}

Typically, single factors identified in the Human and LLama RFRs are similar to multiple factors CFR factors. One rarely sees a one-to-one matching of a Human or Llama RFR factor and a CFR factor.  There are two one-to-one Human RFR / CFR factor matches: ``Refusal to Consent to a Search'' to ``Refused Consent'' and ``Smell and Odors'' to ``Masking Agent.''  There are no one-to-one Llama RFR / CFR matches.

As shown in Table~\ref{fig:llama-dias}, the Llama RFR model also produced multiple mismatches: six factors did not match any CFR Factor.  Eight  CFR factors did not match any Llama RFR factor.  In Table~\ref{fig:human-dias}, a single CFR Factor is connected to four  Human RFR Factors.  In Table~\ref{fig:llama-dias}, many single  CFR Factors were connected to from three to six Llama RFR factors. Incidentally, the GPT RFR produced similar results to the Human RFR in Table~\ref{tab:models_settings}. The diagram for the GPT RFR (not shown) showed similarities to the Human RFR. 

The Llama RFR factors were sometimes too specific and other times too abstract as compared to the Human RFR factors (a similar problem affects humans' factor definitions \cite{rempell2022factors}.) For example, some Llama RFR factors were more specific than CFR factors including oddly specific qualifiers such as ``Unusual or Suspicious Behavior in a High Crime Neighborhood", ``Excessive or Unusual Attention to Luggage or Packages", ``Unusual or Suspicious Explanations for Luggage or Packages". Other Llama RFR factors were too broad such as ``Unusual Behavior".  The mismatches in factors and varying levels of generality in the Llama RFR seem to explain the poor quality of RFR identified by Llama and the commensurate poor predictive power of the RFR in Table~\ref{tab:models_settings}.

Almost all factors synthesized in the RFRs bore some relation to the DIAS domain.  It is worth noting that the cases used in the system were verified as on point to the issue of whether an officer had suspicion of drug trafficking. Nevertheless, the human and LLM RFRs (both \texttt{gpt-4-turbo} and \texttt{llama3-70b-8192}), returned some factors that are not present in the  CFR and are not legally meaningful.  For example, all RFRs identified an Officer's Training and Experience as a factor.  Although this language is mentioned frequently in drug interdiction cases, an officer's experience and training is not a factor.  It has to do with the credibility of an officer's testimony, not the substance of what they're saying.  A judge may be more likely to believe a seasoned officer, but can't use an officer's own personal training as a negative factor against the defendant.  Other examples include the duration of the traffic stop (a separate legal issue) or whether the canine was alerted to the presence of drugs (only after the officer determines they have reasonable suspicion will the dog be used).  The LLM's failure to discount frequently appearing language that is  contrary to basic legal knowledge highlights its limitations in this domain.


Given  the need to detect novel factors reported by judges in case texts, it is interesting that our experimental procedure yielded a plausible candidate. The Llama RFR identified the following ``factor'':  ``Use of Disguise or Deception: A driver’s use of disguise or deception, such as wrapping packages in Christmas paper to blend in with innocent motorists, can support reasonable suspicion.”  We subsequently traced this ``factor” to a case in the data, \textit{Commonwealth of Pennsylvania, Appellee v. Randy Jesus Valdivia,  Appellant}, 145 A.3d 1156, which stated, ``Inside the car, there were two large boxes wrapped in Christmas paper in the backseat. Strangely, the packages were unmarked and undamaged, even though they presumably had been on board Valdivia’s flight to Detroit. Drug smugglers, Trooper Hoy added, often wrap drugs in Christmas paper around the holidays in an effort to blend in with innocent motorists.'' Prior to reviewing the LLM’s outputs, we were unaware of this language in the case. We think this ``factor’s'' explicit focus on disguise or deception could reasonably be categorized as a new factor, not on our pre-defined list of DIAS factors, or, at least, as a sub-factor augmenting the existing DIAS CFR factor `Suspicious Luggage'.  Whether it should be so categorized would depend on whether it is a one-off or has been applied by judges in other DIAS cases.

\subsection{Limitations and Conclusions}

We have demonstrated that it is feasible for an LLM, with a human in the loop, to take a set of raw opinions and produce a representation of a legal domain as factors (RQ2). 
Our attempts to fully automate this process using an LLM resulted in weak to moderate performance (RQ1); so far, it performs better with human involvement.  Our methodology shows promise in identifying new factors for pre-defined factor lists (R3).


A limitation of our work is its reliance on both open-source and proprietary models, with prompts tuned specifically for proprietary models and limited tuning applied to open-source alternatives. The knowledge embedded in these models plays a crucial role in the outcomes. All models, including humans, returned factors that were not legally relevant. This underscores the need to incorporate fundamental legal knowledge into model prompting for more accurate and relevant factor discovery.

Eventually, our methodology could suggest factors to consider for legal argument. Advocates may need to make sense of a collection of legal decisions to identify predictive factors that can align  supportive cases in an argument and distinguish non-supportive ones. Also, with large sets of cases in a domain, researchers could employ the methodology to generate preliminary lists of possible factors to use in conceptually organizing the cases or in predicting case outcomes and for comparing factor lists’ predictive accuracy.

\bibliographystyle{vancouver}
\bibliography{AILaw_bk_2024}

\end{document}